\pdfoutput=1
\documentclass[11pt]{article}

\usepackage[final]{acl}

\usepackage{times}
\usepackage{latexsym}
\usepackage{booktabs}
\usepackage{multirow}
 \usepackage{float}

\usepackage[T1]{fontenc}
\usepackage[utf8]{inputenc}

\usepackage{microtype}

\usepackage{inconsolata}
\usepackage{todonotes}

\usepackage{graphicx}

\usepackage{tikz}
\usepackage{xcolor}
\newcommand{\coloredRectangle}[3]{% #1 color, #2 width, #3 height
    \begin{tikzpicture}
        \fill[#1] (0,0) rectangle (#2,#3);
    \end{tikzpicture}%
}

\graphicspath{ {./images/} }

\title{Can Large Language Models Outperform Non-Experts in Poetry Evaluation? A Comparative Study Using the Consensual Assessment Technique}

\author{\textbf{Piotr Sawicki\textsuperscript{1}},
  \textbf{Marek Grze\'{s}\textsuperscript{1}},
  \textbf{Dan Brown\textsuperscript{1,2}},
  \textbf{Fabr\'{\i}cio G\'oes\textsuperscript{3}}
\\
  \textsuperscript{1}School of Computing, University of Kent, Canterbury, UK \\
  \textsuperscript{2}Cheriton School of Computer Science, University of Waterloo, Canada \\
  \textsuperscript{3}Computing and Mathematical Sciences Department, University of Leicester, UK \\
  \small{
    \texttt{p.sawicki@kent.ac.uk, m.grzes@kent.ac.uk, dan.brown@uwaterloo.ca, fabricio.goes@leicester.ac.uk}
  }
}

\begin{document}
\maketitle
\begin{abstract}
%The Consensual Assessment Technique (CAT) evaluates creativity through holistic expert judgments. We investigate the use of two advanced Large Language Models (LLMs), Claude-3-Opus and GPT-4o, to evaluate poetry by a methodology inspired by the CAT. Using a 90-poem dataset with a ground truth based on publication venue, we found that these LLMs significantly surpass non-expert human judges, especially when evaluating poems in small, randomized batches. Claude-3-Opus exhibited slightly superior performance to GPT-4o. We show that LLMs are viable tools for accurately assessing poetry, paving the way for their broader application in other creative domains.

This study adapts the Consensual Assessment Technique (CAT) for Large Language Models (LLMs), introducing a novel methodology for poetry evaluation. Using a 90-poem dataset with a ground truth based on publication venue, we demonstrate that this approach allows LLMs to significantly surpass the performance of non-expert human judges. Our method, which leverages forced-choice ranking within small, randomized batches, enabled Claude-3-Opus to achieve a Spearman's Rank Correlation of 0.87 with the ground truth, dramatically outperforming the best human non-expert evaluation (SRC = 0.38). The LLM assessments also exhibited high inter-rater reliability, underscoring the methodology's robustness. These findings establish that LLMs, when guided by a comparative framework, can be effective and reliable tools for assessing poetry, paving the way for their broader application in other creative domains.
\end{abstract}

\section{Introduction} \label{section:Introdution}

Quantification of creativity is a complex subject, prompting extensive scholarly debate and resulting in numerous theoretical frameworks for its measurement (e.g.,~\citealp{colton2008creativity,jordanous2012evaluating,jordanous2016four,ritchie2007some,lamb2018evaluating}). While a prevailing strategy within these frameworks involves decomposing creativity into discernible elements, a contrasting approach is the Consensual Assessment Technique (CAT), which offers a holistic evaluation based on the consensus of expert judges~\citep{amabile1983consensual}. Our investigation explores using Claude-3-Opus~\citep{claude3} and GPT-4o~\citep{gpt4o}---two of the best language models at the time of writing---to evaluate poetry through a methodology inspired by the CAT.

Evaluating creativity via human judgment is costly, time-consuming, and can be unreliable with non-experts. This was demonstrated by~\citet{lamb2015human}, who found that crowdsourced judges inverted the ground-truth quality of a 90-poem dataset, ranking professionally published poems lowest and amateur poems highest. More recently, LLMs have been shown to be effective evaluators in other domains; for instance, research on using GPT-4 for chatbot responses found that its judgments achieved over 80\% agreement with the consensus of human raters~\citep{zheng2024judging}. This demonstrated capacity for nuanced assessment motivates our re-evaluation of the~\citet{lamb2015human} dataset using Claude-3-Opus and GPT-4o, in order to establish whether automated evaluation can surpass the previously documented performance of non-expert human assessments.

Our primary contribution is the introduction of a methodology for evaluating poetry using LLMs that is inspired by the CAT. We demonstrate that LLM evaluations exhibit strong correlations with the ground truth and significantly outperform non-expert human judges in accurately categorizing poems based on their quality, while maintaining strong inter-rater reliability between consecutive executions. These results indicate that the use of LLMs in accurately assessing human creative works is becoming increasingly feasible and promising.

The subsequent sections of the paper are structured as follows: The Related Work section reviews prior work on creative evaluation and batch-wise evaluation. Next, the Methodology section details our CAT-inspired approach, followed by a description of the Dataset and the benchmark Results of Human Evaluation. We then present our four main experiments: Experiment 1 explores without-context poem classification; Experiments 2 and 3 investigate in-context evaluation using prompts of 90 and 15 poems, respectively; and Experiment 4 assesses the inter-rater reliability of the LLMs. Finally, in the Discussion and Conclusion, we summarize key findings and propose future research directions.

\section{Related Work} \label{section:Related Work}

Our research builds on and contributes to two main areas: the use of LLMs for creative evaluation and the emerging paradigm of batch-wise evaluation for improved performance and efficiency.

\subsection{LLMs as Creative Evaluators}\label{subsection:LLMs as Creative Evaluators}

Research into the capacity of LLMs to emulate human judgment in creative domains has progressed rapidly. Early studies by~\citet{goes2022crowd} explored GPT-3.5’s performance in evaluating comedic content, with subsequent advancements focusing on GPT-4’s enhanced capabilities~\citep{goes2023gpt}. In the domain of creative writing,~\citet{chakrabarty2024art} found that while LLMs could generate and evaluate short stories, they were less effective at evaluation than human experts, underscoring the challenges of assessing nuanced creative content. Poetry, as a distinct and compact creative form, has also been a subject of study. Our previous work has explored using LLMs as binary classifiers to distinguish human poems from AI poems generated to replicate a specific author's style~\citep{sawicki2022training, sawicki2023power,sawicki2023bits}. In contrast,~\citet{agirrezabal2023erato} has focused on evaluating structural aspects like rhyme and meter. In their computational analysis of poetry,~\citet{kao-jurafsky-2015-computational} found objective stylistic differences that distinguish professional and amateur work, showing the divide is not merely subjective: contemporary professional poems—heavily influenced by the Imagist movement—feature higher word concreteness and more subdued emotional language, while contemporary amateur poems stylistically resemble older, 19th-century professional works. This linguistic analysis provides a strong justification for the ground truth used in our study, which is based on the prestige of publication venues, by showing that such distinctions correspond to measurable features.

\subsection{Batch-wise and In-Context Evaluation}\label{subsection:Batch-wise and In-Context Evaluation}

The standard method for LLM-based assessment has been ``sample-wise'', where each item is evaluated in a separate prompt. However, our in-context evaluation of multiple poems aligns with the emerging paradigm of ``batch evaluation'', which leverages inter-sample comparison to improve judgments.
The paradigm of batch evaluation for improving LLM judgments has been formalized for tasks like dialogue and summarization in frameworks such as BatchEval~\citep{yuan-etal-2024-batcheval}. Their method involves an iterative process, creating ``heterogeneous'' batches to enhance the evaluator's robustness and accuracy. Although our work also leverages in-batch comparison, our methodology differs: we use a simple, non-iterative approach with randomized subsets, guided by forced-choice evaluation principles~\citep{brown2018modelling}, and apply it to the novel domain of poetry.

Other work has framed batching as an efficiency tool, warning that evaluation quality can degrade with larger batch sizes~\citep{larionov2025batchgembatokenefficientmachinetranslation}. Our work, however, reframes the role of batching in the creative domain. We demonstrate that reducing the batch size from 90 to 15 poems is not a trade-off but a direct mechanism for enhancing accuracy. This method of forced-choice comparison within smaller batches significantly improves correlation with the ground truth over single-item or large-batch approaches, suggesting that for creative assessments, batch design is a important factor for achieving reliable judgments.

\section{Methodology} \label{section:Methodology}

The Consensual Assessment Technique (CAT)~\cite{amabile1983consensual} is a method for evaluating creative products by asking experts in the domain to provide their own holistic judgments of creativity, without being given explicit criteria to follow. Research confirms CAT’s reliability~\citep{baer2009assessing}, provided the judges have practical domain expertise~\citep{kaufman2009expertise}. Key protocols include independent evaluations, creator anonymity, and the use of a numerical rating scale. While our study does not use a panel of human experts and thus does not strictly replicate the CAT, our methodology is inspired by its core principle of holistic, comparative assessment.

Our adaptation operationalizes CAT's philosophy by leveraging LLMs in four specific ways. First, we treat Claude-3-Opus and GPT-4o as \textbf{surrogate judges}, leveraging their extensive training on vast corpora of human text to approximate domain understanding. Second, we implement CAT's principle of relative judgment through an \textbf{in-context evaluation} approach, where poems are assessed in randomized batches. Third, we \textbf{compel comparative judgments} by instructing the LLMs to produce a ranked list, a method that aligns with forced-choice evaluation and can increase differentiation between items~\citep{brown2018modelling}. Finally, we \textbf{simulate a panel of independent raters} by setting the \texttt{temperature} hyperparameter to 1 for all queries. Temperature regulates sampling randomness~\citep{holtzman2020curious}, and this setting introduces variability analogous to the differing opinions of individual experts, while also allowing for an analysis of inter-rater reliability. The practical effect of this variability is demonstrated in Experiment 4 (see Table~\ref{tab:poetry_scores_example_iccc} in Appendix~\ref{app:repeated_evaluations_example}).

Our framework instructs the LLMs to appraise poems against five distinct criteria: Creativity, Quality, Innovativeness, Similarity, and Poeticness. This set was carefully adapted from the nine criteria used by~\citet{lamb2015human} for two key reasons. First, the original study reported very high correlations between several of its criteria---most notably between ``Skill'' and ``Appreciation'' ($r=0.99$)---suggesting significant conceptual overlap and experimental redundancy. Second, two of their criteria were inherently anthropocentric. The IDEA model's criteria of ``Wellbeing'' (the reader's enjoyment) and ``Effort'' (willingness to engage) measure a subjective, internal human experience that an LLM can only simulate, not genuinely possess. By excluding these and consolidating others, our chosen criteria focus squarely on the assessable properties of the creative artifact itself. Our ``Quality'' criterion consolidates their ``Quality'' and ``Value'', which were measured with identical items. Our ``Innovativeness'' and ``Similarity'' are direct counterparts to their ``Novelty'' and ``Typicality''. The broader ``Creativity'' criterion was introduced in order to capture the essence of what Lamb et al. measured through multiple, highly-correlated criteria (e.g., ``Imagination'', ``Skill''). Finally, ``Poeticness'' serves as a foundational check, inspired by a reverse-coded item for ``Typicality'' in the original study (``This is not a poem''). To implement this, we left the definitions of ``Creativity'' and ``Quality'' open to the LLMs' holistic interpretation. In contrast, ``Innovativeness'' was explicitly defined as ``this poem is not like other poems I have seen before,'' with ``Similarity'' as its reverse. ``Poeticness'' served to assess whether a text qualifies as a poem. The exact prompts are in Appendix~\ref{section:5_criteria_prompts}.

The specific models utilized in our evaluations were the 2024-02-29 version of Claude-3-Opus and the 2024-05-13 version of GPT-4o.

\section{Dataset} \label{section:Dataset}

\citet{lamb2015human} present a dataset comprising 90 poems divided into three distinct categories: ``Good,'' ``Medium,'' and ``Bad''. This categorization serves as a robust ground truth. Poems classified as ``Good'' were sourced from the magazine \textit{Poetry}, regarded as the foremost English language poetry journal globally. Those categorized as ``Medium'' were obtained from intermediate level poetry magazines that offer remuneration of \$5-\$10 per poem. Poems in the ``Bad'' category were selected from the Newbie Stretching Room at the Poetry Free-For-All website~\citep{PFFA}, and the authors specifically chose  works with no positive feedback. Submissions to \textit{Poetry} magazine are subjected to meticulous scrutiny by its editorial team. Similarly, poems published in other poetry magazines undergo a degree of editorial review. In contrast, the Newbie Stretching Room permits unrestricted publication, indicating a lack of editorial filtration. Consequently, it is the publication venue that establishes the ground truth for this experiment. While copyright issues prevent us from reproducing the full dataset, the poems from the ``Good'' category are publicly archived online. We provide direct links to these poems in Appendix~\ref{llinks} to offer readers concrete examples from the corpus. To further aid reproducibility, the full dataset will be made available to researchers upon request.

While it is true that poetry is inherently subjective, the categorization of poems based on their publication venue is analogous to the evaluation of research papers in science by venue, where papers published in prestigious venues are generally considered to be of high quality, while those published in mid-level venues are usually regarded as being of medium quality, and papers published without any peer review have traditionally been viewed as being of poor quality. This hierarchy of publication venues, a common practice in scientometrics, serves as a proxy for the quality of the work~\citep{hicks2015bibliometrics, franceschet2010role}, much like the categorization of poems in this dataset. Moreover, just as some poems win competitions and are included in prestigious anthologies, some research papers receive awards and are widely cited within their respective fields. Therefore, while the subjective nature of poetry evaluation cannot be denied, the use of publication venue as a ground truth for categorizing poems is a reasonable approach.

\begin{figure*}[!ht]
\begin{center}
% Adjust the rectangles' width and height as needed
\newcommand{\rectWidth}{0.08} % Width in cm
\newcommand{\rectHeight}{0.3} % Height in cm
\small
\setlength{\tabcolsep}{4pt}
\begin{tabular}{@{}lcr@{}}
\toprule
Criterion & Ordering & SRC  \\
\bottomrule
\end{tabular}%
\vspace{2mm}
\end{center}
\caption{The poem orderings obtained in evaluations by human non-expert judges, using data from~\citet{lamb2015human}. Each poem is colour-coded by its ground truth category, and for each criterion, poems are ranked from highest-rated (left) to lowest-rated (right). The figure highlights a significant inversion of the ground truth for most criteria; for example, the predominance of red-coded (``Bad'') poems at the top of the ``Typicality'' ranking demonstrates that judges rated the lowest-quality poems most highly.}
\label{fig:ordering_human_only}
\end{figure*}

The categories were encoded as follows: ``Good'' = A, ``Medium'' = B, and ``Bad'' = C. Each category comprises 30 poems, resulting in the ground truth order: [A ... (30 times), B ... (30 times), C ... (30 times)]. In rank correlation used to analyze our results, A corresponds to rank 1, B to rank 2, and C to rank 3 (see Figures~\ref{fig:ordering_human_only} and~\ref{fig:ordering_claude15n},  and Figure~\ref{fig:ordering_all4_human} in the Appendix~\ref{app:Visual Representation of Poem Orderings}).

 It should be noted that although the ground truth consists of three distinct groups which are ordered by quality, there is no ground truth ranking of the poems within the groups. Therefore, we will study how well the rankings returned by the LLMs identify the categories the poems belong to.

\section{Results of Human Evaluation} \label{section:Results of Human Evaluation}

The benchmark for our study is the non-expert human evaluation conducted by \citet{lamb2015human}. Their findings revealed a striking inversion of the ground truth: non-expert judges consistently rated poems from the “Good” category as the lowest quality and those from the “Bad” category as the highest. While the precise reasons for this inversion are complex, we might speculate that this outcome is rooted in stylistic evolution. Drawing on the work of \citet{kao-jurafsky-2015-computational}, it is plausible that the non-expert judges found the stylistic qualities of the ``Bad'' poems—which often resemble more traditional, 19th-century professional works—to be more familiar and accessible. Conversely, the ``Good'' poems, representing the forefront of contemporary professional poetry, may exhibit more modern, perhaps Imagist-influenced sensibilities. It is conceivable that these qualities, while valued by experts, were perceived as less engaging or more opaque by a non-expert audience, leading to their lower ratings.

\begin{table}[htbp] % Added placement specifier
\centering % Changed from \begin{center}
\small
\setlength{\tabcolsep}{8pt} % Retained custom tabcolsep
{
\begin{tabular}{l r r} % Removed vertical lines, 'r' for numerical columns
\toprule
\textbf{Criterion} & \textbf{SRC}   & \textbf{p-value} \\
\midrule
\textbf{Novelty}   &  \textbf{0.38} & \textbf{1.92e-04}  \\
Imagination        & -0.12          & 0.27      \\
Value              & -0.20& 0.06      \\
Quality            & -0.33          & 1.32e-03  \\
Appreciation       & -0.33          & 1.33e-03  \\
Skill              & -0.35          & 7.21e-04  \\
Typicality         & -0.37          & 3.79e-04  \\
Wellbeing          & -0.45          & 8.59e-06  \\
Effort             & -0.45          & 8.59e-06  \\
\bottomrule
\end{tabular}
}
%\caption{Spearman's Rank Correlation (SRC) results from the non-expert human evaluations conducted by~\citet{lamb2015human}. The data reveals a striking inversion of the ground truth, with most criteria showing a significant negative correlation, indicating that judges preferred lower-quality poems. The sole exception, ``Novelty'' (SRC = 0.38), represents the best-performing human judgment and serves as the primary benchmark for our LLM evaluations.}\label{tbl:human_results}
\caption{Spearman's Rank Correlation (SRC) results from the non-expert human evaluations conducted by~\citet{lamb2015human}. The data shows a significant negative correlation for most criteria, with ``Novelty'' being the sole exception.}\label{tbl:human_results}
\end{table}

%The Spearman's rank correlations, presented in Table~\ref{tbl:human_results}, quantify this reversal. Most criteria, such as ``Quality'' (SRC = -0.33) and ``Skill'' (SRC = -0.35), were significantly and negatively correlated with the quality-by-venue ground truth. In fact, the strongest negative correlations were for ``Wellbeing'' and ``Effort'' (both SRC = -0.45), indicating that non-experts most enjoyed and were willing to engage with the poems of the lowest quality. The sole exception was ``Novelty,'' which achieved a positive correlation (SRC = 0.38). This result stands as the best-performing non-expert human evaluation and serves as the primary benchmark for our LLM comparisons. The subsequent sections will show that LLM-based evaluations can progressively and significantly surpass this benchmark.

The Spearman's rank correlations in Table~\ref{tbl:human_results} quantify this reversal, which is visualized by the poem orderings in Figure~\ref{fig:ordering_human_only}. Most criteria, such as ``Quality'' (SRC = -0.33) and ``Skill'' (SRC = -0.35), were significantly and negatively correlated with the quality-by-venue ground truth. In fact, the strongest negative correlations were for ``Wellbeing'' and ``Effort'' (both SRC = -0.45), indicating that non-experts most enjoyed and were willing to engage with the poems of the lowest quality. The sole exception was ``Novelty,'' which achieved a positive correlation (SRC = 0.38). This result stands as the best-performing non-expert human evaluation and serves as the primary benchmark for our LLM comparisons. The subsequent sections
\begin{table*}[ht]
\centering 
\small
{

\begin{tabular}{l rrrrrrr} 
\toprule
\textbf{Model} & \textbf{Good} & \textbf{Medium} & \textbf{Bad} & \textbf{Total} & \textbf{Accuracy} & \textbf{SRC} & \textbf{p-value} \\
\midrule
\textbf{GPT-4o-2024-05-13} & \textbf{37 (21)} & \textbf{42 (15)} & \textbf{11 (11)} & \textbf{90 (47)} & \textbf{52.2\%} & \textbf{0.62} & \textbf{5.55e-11} \\
Claude-3-Opus          & 23 (16)          & 47 (22)          & 20 (16)          & 90 (54)          & 60\%            & 0.57          & 3.77e-9  \\
GPT-4-2024-04-09       & 42 (22)          & 47 (11)          & 1 (1)            & 90 (34)          & 37.8\%          & 0.57          & 2.55e-9  \\
GPT-4-2023-11-16       & 32 (15)          & 58 (16)          & 0 (0)            & 90 (31)          & 34.4\%          & 0.34          & 1e-3     \\
\bottomrule
\end{tabular}
}
\caption{Results of the without-context classification task from Experiment 1, where each poem was classified individually. The columns ``Good'', ``Medium'', and ``Bad'' show the number of poems assigned to each category, with the number of correct predictions in parentheses. Notably, all recent models significantly surpassed the best non-expert human evaluation (SRC = 0.38, see Table~\ref{tbl:human_results}), with GPT-4o achieving the highest Spearman's Rank Correlation of 0.62. The data also illustrates a performance improvement in newer model versions and a tendency to favor the ``Medium'' category.}\label{tbl:3_cat_results}
\end{table*}
\noindent will show that LLM-based evaluations can progressively and significantly surpass this benchmark.

\section{Experiment 1---Evaluating Poems Without Context} \label{section:Experiment 1}

While the primary focus of this paper is on in-context evaluations, we have also conducted a simple without-context evaluation where we prompted the LLMs to classify each poem into its publishing category using the prompt in Appendix~\ref{section:GPT3-ABC}. To gain insight into the evolution of GPT-4 models, we tested not only GPT-4o but also two older versions: GPT-4-2024-04-09 and GPT-4-2023-11-06. After classifying the poems, we produced an ordered list of poem categories and compared it to the ground truth using Spearman's rank correlation (with ranks A, B and C). The results are presented in Table~\ref{tbl:3_cat_results}. Three out of four models tested (Claude-3-Opus, GPT-4o-2024-05-13 and GPT-4-2024-04-09) have already exceeded the best human evaluation results by a significant margin.

The highest Spearman rank correlation of 0.62 between its rank order and the ground truth was achieved by GPT-4o, followed by Claude-3-Opus and GPT-4-2024-04-09 (SRC=0.57). These exceed the 0.38 obtained by non-expert humans assessing Novelty. The distribution of poem categories was uneven in all four of the evaluated models, an issue exacerbated in older versions of GPT-4. The results show a clear trend where the accuracy of category distribution improves with newer versions. We suspect that the uneven distribution, particularly the preference for the middle category, is the result of the models' training bias, which may encourage moderation to avoid errors and extreme judgments, while avoiding negative feedback.  

While single-item classification has already surpassed the non-expert human baseline, in the following experiments, the models are required to rank poems within a set, an approach designed to leverage comparative context for even more precise evaluations.

\section{In-context Poem Evaluation} \label{secion:In Context}

We first explain how our prompts (Appendix~\ref{section:5_criteria_prompts}) for in-context evaluation force the LLMs towards comparative assessments. To this end, the prompts ask the LLMs to evaluate every poem in a set on a scale $1-5$, as these values worked best in our exploratory experiments. Additionally, the prompts (Appendix~\ref{section:5_criteria_prompts}) ask the LLMs to return the ordered list of poems. This second requirement brings us even closer to the forced-choice methods \cite{brown2018modelling}. These two requirements in the prompts afford two scoring methods based on the same LLM outputs:
\begin{enumerate}
    \item Evaluations extracted from the $1-5$ scores will be referred to as Claude-3-Opus 90, Claude-3-Opus 15 and GPT-4o 15, where numbers \textit{90} and \textit{15} indicate the number of poems in a prompt.
    \item The poem's rank in the LLM's output list can be used as a score, where the first poem receives a score of 15, the second 14, and so on, until the last poem receives a score of 1. Evaluations done it this way will be referred to as Claude-3-Opus 90n, Claude-3-Opus 15n and GPT-4o 15n, where the numbers \textit{90} and \textit{15} indicate the number of poems in a prompt.
\end{enumerate}
Note that the two scoring methods presented above yield the same order for one set of poems because they are part of the same output. However, when the averaged poem scores over 100 sets are computed, the two methods induce different orderings of the poems.

\subsection{Analyzing Input Size for In-context Poem Evaluation} \label{subsection:Analyzing Input Size}

Claude-3-Opus can analyze all 90 poems in a single prompt. In the initial experiments, 16 out of 20 such attempts were successful (i.e. the ranking of all 90 poems is provided in the LLM response). Unsuccessful responses contained an incomplete list of poems, or contained duplicates, or the LLM responded that it is incapable of fulfilling this task. 

For GPT-4o, only 3 out of 20 attempts to rank 90 poems were successful. This could be because of the phenomenon observed in~\citet{liu2024lost}---while LLMs are capable of accepting very large prompts, the performance decreases with the length of the prompt, and the models do not handle equally well the whole content of the prompt. As a result, we reduced the GPT-4o input to 30 poems per prompt, resulting in a successful analysis in 17 out of 20 attempts. Further reduction to 15 poems per prompt increased the success rate to 100\%. Claude-3-Opus also successfully ranked the 15-poem prompts on every attempt.

\begin{table*}[ht]
\centering
\small
\setlength{\tabcolsep}{5.5pt}
\begin{tabular}{@{}l rrrrrrrrrrrr@{}}
\toprule
& \multicolumn{2}{c}{\textbf{Claude-3 90n}} & \multicolumn{2}{c}{\textbf{Claude-3 90}} & \multicolumn{2}{c}{\textbf{Claude-3 15n}} & \multicolumn{2}{c}{\textbf{Claude-3 15}} & \multicolumn{2}{c}{\textbf{GPT-4o 15n}} & \multicolumn{2}{c}{\textbf{GPT-4o 15}} \\
\cmidrule(lr){2-3} \cmidrule(lr){4-5} \cmidrule(lr){6-7} \cmidrule(lr){8-9} \cmidrule(lr){10-11} \cmidrule(lr){12-13}
\textbf{Criterion} & \textbf{SRC} & \textbf{p-val} & \textbf{SRC} & \textbf{p-val} & \textbf{SRC} & \textbf{p-val} & \textbf{SRC} & \textbf{p-val} & \textbf{SRC} & \textbf{p-val} & \textbf{SRC} & \textbf{p-val} \\
\midrule
Innovativ. & 0.63 & 2.1e-11 & \underline{0.68} & 1.2e-13 & 0.70 & 1.6e-14 & 0.67 & 7.4e-13 & \textbf{0.75} & 1.8e-17 & 0.73 & 2.0e-16 \\
Poeticness & \underline{0.67} & 7.36e-13 & 0.63 & 2.1e-11 & \textbf{0.80} & 3.2e-21 & 0.78 & 7.2e-20 & 0.68 & 1.2e-13 & 0.70 & 1.6e-14 \\
Creativity & 0.63 & 2.1e-11 & 0.62 & 9.7e-11 & 0.72 & 2.0e-15 & 0.68 & 1.2e-13 & 0.72 & 2.0e-15 & \textbf{\underline{0.77}} & 1.3e-18 \\
Quality & 0.57 & 5.8e-09 & 0.57 & 5.8e-09 & \textbf{\underline{0.87}} & 2.6e-28 & \underline{0.85} & 3.2e-26 & \underline{0.82} & 1.0e-22 & 0.77 & 1.3e-18 \\
Similarity & 0.47 & 3.5e-06 & 0.52 & 1.9e-07 & \textbf{0.83} & 2.2e-24 & 0.80 & 3.2e-21 & 0.70 & 1.6e-14 & 0.72 & 2.0e-15 \\
\bottomrule
\end{tabular}
\caption{Spearman's Rank Correlation (SRC) results for all in-context evaluations from Experiments 2 (90-poem prompts) and 3 (15-poem prompts). All methods dramatically outperform the best non-expert human benchmark (SRC = 0.38, Table~\ref{tbl:human_results}). The results highlight two key findings: (1) evaluating smaller batches of 15 poems yields significantly higher correlations than evaluating all 90 at once, and (2) deriving scores from the poem's rank (``n'' methods) generally improves performance, with Claude-3-Opus 15n achieving the highest overall correlation of 0.87 on the ``Quality'' criterion. The highest SRC for each criterion (row) is \textbf{bolded}, and the best result for each model/method (column) is \underline{underlined}.}
\label{tbl:exp_1_90_correlations}
\end{table*}

Therefore, we will present the evaluations of 90 poems in one prompt for Claude-3-Opus only, and evaluations relying on 15 poems in one prompt for both Claude-3-Opus and GPT-4o.

\subsection{Experiment 2---Evaluation of Poems---90 poems in a prompt} \label{subsection:Experiment 2}

Since the GPT-4o version used in this study was incapable of evaluating all 90 poems in a single prompt, this experiment is conducted only with Claude-3-Opus. For prompts, we have prepared 10 datasets containing all 90 poems, where the poems' ordering for each dataset was randomized. Thus, each poem will be evaluated ten times.

Each of the ten 90-poem datasets underwent evaluation in the specified five criteria: Creativity, Quality, Innovativeness, Similarity, and Poeticness, using the prompts presented in Appendix~\ref{section:5_criteria_prompts}. Post-evaluation, the numerical scores for each poem were averaged, and the poems were ranked based on their average scores. We have used both the original $1-5$ scale scores assigned by the model (Claude-3-Opus 90), and the poem's positions on the list (Claude-3-Opus 90n).

After sorting the poems by their average evaluation score, each poem's position was marked with the category it represents (A, B or C), thus producing an ordered list of A's, B's and C's. These lists were then compared to the ground truth, and  Spearman’s rank correlation was computed, using the ranks A, B, and C. Table~\ref{tbl:exp_1_90_correlations} presents Spearman's correlation coefficients calculated against the ground truth order, and Figure~\ref{fig:ordering_all4_human} (Appendix) presents graphical representations of the results. These outcomes demonstrate an even higher degree of correlation with the ground truth (SRC=0.68) as compared to ``without-context'' evaluations, and are all statistically significant, with very low p-values in every case. Unsuccessful attempts, which resulted in incomplete or malformed output, were discarded, and the prompt was re-submitted until a valid response was generated. Subsequent experiments will show that reducing the number of poems in the prompt improves the accuracy even further.

\subsection{Experiment 3---Evaluation of Poems---15 poems in a prompt}
\label{subsection:Experiment 3}

In this experiment, 5 poems were randomly selected from each category (A, B and C) to form a subset of 15 poems for each query to both LLMs. This process was repeated to create 100 unique subsets, each containing poems from all three categories and shuffled accordingly.

\begin{figure*}[!ht]
\begin{center}
% Adjust the rectangles' width and height as needed
\newcommand{\rectWidth}{0.08} % Width in cm
\newcommand{\rectHeight}{0.3} % Height in cm
\small
\setlength{\tabcolsep}{4pt}
\begin{tabular}{@{}lcr@{}}
\toprule
Criterion & Ordering & SRC  \\
\bottomrule

\end{tabular}%
\end{center}
\caption{Poem orderings generated by Claude-3-Opus 15n, the best-performing model and method from Experiment 3 (Section \ref{subsection:Experiment 3}). The results demonstrate a strong correlation with the ground truth, with the ``Quality'' criterion achieving a Spearman's Rank Correlation (SRC) of 0.87—a significant improvement over the best human non-expert evaluation (SRC = 0.38, see Figure \ref{fig:ordering_human_only}).}
\label{fig:ordering_claude15n}
\end{figure*}

Each of the 100 15-poem datasets underwent evaluation using the same five criteria: Creativity, Quality, Innovativeness, Similarity, and Poeticness, using the prompts shown in Appendix~\ref{section:5_criteria_prompts}.

Due to the random sampling across 100 subsets, the frequency of individual poems varied slightly. This selection process can be modeled as a binomial distribution where the number of trials $n$ is 100 (the number of subsets) and the probability $p$ of any single poem being selected is 5/30. The expected number of appearances, $\mu$, for any poem is given by the formula:
\begin{equation}
\mu = n \times p
\end{equation}
This yields an expected frequency of $100 \times (5/30) \approx 16.7$ appearances. In our experiment, the actual number of appearances for any given poem fell within the statistically expected range of 9 to 25 times. Because each poem's final score is an average taken across numerous, independently randomized contexts, this minor variation in appearance frequency is not expected to introduce systematic bias into the results. After evaluation, scores for each poem were averaged across all subsets where they appeared, and the poems were ranked by their average evaluation scores. This was done for both types of scoring methods.

As before, each poem's position in the scoring list was marked with its category (A, B or C), resulting in an ordered list. These lists were compared to the ground truth using Spearman’s rank correlation with ranks A, B and C. Table~\ref{tbl:exp_1_90_correlations} presents those correlations, showing a very high degree of consistency with the ground truth, and all values are statistically significant with extremely low p-values.

Overall, most results surpass those obtained by Claude-3-Opus when evaluating all 90 poems in a single prompt and significantly surpass those of human non-expert judges. Evaluations of 15 poems per prompt yielded much higher correlations to the ground truth. The highest correlation, 0.87, was achieved in the Quality criterion by Claude-3-Opus (Table~\ref{tbl:exp_1_90_correlations} and Figure~\ref{fig:ordering_claude15n} and Figure~\ref{fig:ordering_all4_human} in Appendix~\ref{app:Visual Representation of Poem Orderings}) and  when the scores were the poem's positions in the list. For GPT-4o, the results were slightly lower, but the Quality criterion also yielded the highest correlation when the scores were the poem's positions in the list. While not universally superior across all criteria, the rank-based scoring method ('n') produced the highest overall correlation with the ground truth (SRC = 0.87 with Claude-3-Opus 15n). For both Claude-3-Opus and GPT-4o, this rank-based method also led to a stronger separation between the quality tiers, a point reinforced by the higher F-values observed in the subsequent ANOVA analysis (see Section~\ref{sec:anova}).

\subsubsection{Analysis of Variance}\label{sec:anova}

Similar to~\citet{lamb2015human}, we conducted a single-factor ANOVA (Analysis of Variance) for each evaluation criterion. ANOVA is a statistical test used to determine whether there are any statistically significant differences between the mean scores of three or more independent groups. In our case, we used it to compare the average evaluation scores across the “Good,” “Medium,” and “Bad” poem categories. For each of our five criteria, the null hypothesis stated that the mean scores for all three poem categories are equal, while the alternative hypothesis suggested that at least one category mean is different.

\begin{table}[htbp]
\centering
\small
\setlength{\tabcolsep}{4pt}
\begin{tabular}{@{}l rrrrr@{}}
\toprule
\textbf{Criterion} & \textbf{Good} & \textbf{Medium} & \textbf{Bad} & \textbf{F-value} & \textbf{p-value} \\
\midrule
Innovativeness & 10.83 & 9.02 & 4.26 & 52.89 & 9.3e-16 \\
Quality        & 11.65 & 9.06 & 3.33 & 132.23& 4.2e-27 \\
Creativity     & 11.11 & 8.95 & 4.02 & 62.55 & 1.46e-17\\
Poeticness     & 11.42 & 8.81 & 3.79 & 85.15 & 3.27e-21 \\
Similarity     & 11.22 & 8.63 & 4.13 & 111.08& 1.11e-24 \\
\bottomrule
\end{tabular}
\caption{Analysis of Variance (ANOVA) for our best-performing method, Claude-3-Opus 15n. The distinct separation of mean scores in the expected order (Good > Medium > Bad) is statistically significant across all criteria, as confirmed by the high F-values and extremely low p-values (p << 0.01). This demonstrates the model's ability to reliably differentiate between the ground truth quality tiers.}\label{tbl:var_1_results}
\end{table}

The results, presented in Table~\ref{tbl:var_1_results} and Table  \ref{tbl:var_1_results_full} in Appendix~\ref{ANOVA results for poem evaluation in five criteria in Experiments 2 and 3}, indicate statistically significant differences between the poem category means for each criterion, considering the Bonferroni-corrected alpha level of 0.01. This finding implies that there are significant variations among the groups for all tested criteria, suggesting that a poem's category has a measurable impact on the scores of all five criteria.

We can also observe that using the poem's position in the output list as a scoring method increases the F-value and decreases the p-value in every case, indicating higher effect size of this approach.

\section{Experiment 4---Evaluating Poetry - Interrater Reliability of Claude-3-Opus and GPT-4o} \label{subsection:Experiment 4}

The primary goal of the previous experiments was to approximate the ground truth categorization. Now, we will test if the LLMs return different rankings when the same set-up is repeated several times (the LLM's temperature parameter was set to 1 to maximize randomness in all our experiments in this paper). After that, we check if the diverse outputs are consistent and reliable with respect to the rankings they yield. For that we conducted a dedicated experiment where we have randomly selected 1 subset of 15 poems (from the subsets used in Experiment 3) and evaluated this subset 10 times with our LLMs. We are evaluating both models with both scoring methods. Effectively, 200 evaluations were conducted (2 models $\times$ 2 ways of scoring  $\times$ 10 repetitions $\times$ 5 criteria).

In Table~\ref{tab:poetry_scores_example_iccc} in Appendix~\ref{app:repeated_evaluations_example} we present an example of evaluating the ``Creativity'' criterion with Claude-3-Opus 15n and Claude-3-Opus, where the outputs are different, yet similar, across multiple runs of the same set-up.

To gauge the reliability of these diverse LLM outputs, we calculated the Intraclass Correlation Coefficient (ICC)~\citep{shrout1979intraclass}. The ICC is a descriptive statistic that measures the consistency or agreement of ratings among multiple judges—in our case, the 10 repeated evaluation runs of an LLM—when assessing the same set of items. A high ICC value (typically > 0.7) indicates strong reliability. There are several ICC models, with the "k-rater" variants (i.e., ICC1k, ICC2k and ICC3k) being most appropriate for our design, as they assess the reliability of averaged ratings from multiple runs. In Table~\ref{tab:ICC_full} (Appendix~\ref{app:Results of ICC evaluations}), we present the results for all ICCk evaluations.

The correlation results of ICC1k, ICC2k and ICC3k were all very similar, and they ranged between 0.9 and 0.99 for evaluations of in all 5 criteria. In every case, using the scores derived from the poem's position in the output list has significantly lowered the p-values and increased the F-scores. All ICCk tests show very high correlation between raters and all results are statistically significant. 

Overall, these results confirm that while the temperature parameter successfully diversifies the output of LLMs (Table~\ref{tab:poetry_scores_example_iccc} in Appendix~\ref{app:repeated_evaluations_example}), the ICC results show that those diversified outputs are stable and reliable on average (Table~\ref{tab:ICC_full} in the Appendix~\ref{app:Results of ICC evaluations}).

\section{Discussion} \label{section:Discussion}

The results of our experiments demonstrate a clear progression in the evaluative capabilities of LLMs, consistently and significantly surpassing the performance of non-expert human judges. Even the simplest, without-context classification (Experiment 1) yielded a Spearman's Rank Correlation of up to 0.62 (GPT-4o), already a substantial improvement over the best human baseline (SRC = 0.38). The shift to an in-context, comparative methodology further enhanced performance. Evaluating the entire 90-poem dataset in a single prompt (Experiment 2) pushed the correlation as high as 0.68. The most significant leap, however, came from reducing the input size: evaluating small, randomized 15-poem subsets (Experiment 3) allowed Claude-3-Opus to achieve an SRC of 0.87, confirming that a forced-choice ranking within smaller batches is a highly effective method for this task.

The superiority of the 15-poem, rank-based scoring method can be attributed to two factors. First, the use of smaller batches likely reduces the cognitive load on the models, mitigating the ``lost in the middle'' effect where performance degrades on long contexts, a phenomenon noted in our own initial tests and by others~\citep{liu2024lost}. This allows the LLM to perform a more focused and effective comparison. Second, the rank-based scoring ('n' method) proved more effective than the 1-5 scale because it forces the model to make finer-grained distinctions. By requiring a complete ordered list, it compels a relative judgment for every single poem, preventing the model from clustering multiple poems at the same score level (e.g., assigning a ``3'' to several poems) and thereby producing a richer, more differentiated signal of quality.

Furthermore, the LLM evaluations proved not only accurate but also highly reliable. The Intraclass Correlation Coefficients from Experiment 4, ranging from 0.90 to 0.99, indicate a remarkable level of consistency across repeated runs. These values are particularly noteworthy as they exceed the typical thresholds of 0.7 to 0.9 that are often considered acceptable for panels of human judges in CAT studies. This high scores suggest that the forced-choice methodology is a reliable driver of the models' strong performance. By compelling the LLMs to make nuanced, relative judgments rather than absolute ones, our method enhances their ability to consistently differentiate between the ground-truth quality tiers, resulting in a stable and reproducible assessment tool.

Looking forward, the success of this CAT-inspired framework suggests it is a promising and durable approach for automated creative assessment. As LLM capabilities continue to evolve, a critical next step will be to determine if this methodology can enable them to match the evaluations of domain \textit{experts}, moving beyond the non-expert benchmark surpassed here. However, this potential also raises important questions about aesthetic diversity. The observed consensus between Claude-3-Opus and GPT-4o could reflect shared training data biases. Future work must focus on developing methods to probe and mitigate these biases to avoid inadvertently reinforcing narrow or homogenized notions of literary quality, ensuring that these powerful tools augment, rather than constrain, human creativity.

% This is the new, more concise Conclusion section
\section{Conclusion} \label{section:Conclusion}

Our experiments have shown that Large Language Models, specifically Claude-3-Opus and GPT-4o, can serve as effective evaluators of poetry. When used within our CAT-inspired, in-context framework, they achieve assessments that show strong correlations with ground truth and markedly outperform non-expert human judges. High inter-rater reliability across multiple criteria further underscores the consistency and robustness of this LLM-based evaluation approach.

This study contributes several key insights to the field of automated creative assessment:

\begin{enumerate}
    \item \textbf{Novel Evaluation Method:} We introduce a CAT-inspired methodology showing that LLMs can provide consistent, reliable assessments of creative work across multiple criteria.

    \item \textbf{Optimized Batch Evaluation:} We demonstrate that evaluating smaller, curated subsets of poems significantly enhances assessment accuracy compared to evaluating an entire dataset at once.

\item \textbf{Effective Scoring Methodology:} We find that deriving scores from a poem’s ranked position in an output list is a highly effective scoring methodology, yielding the best overall correlation with the ground truth.
\end{enumerate}

A direct comparison between LLM assessments and scores from a live panel of expert judges remains a crucial next step for validating these findings in professional contexts. As LLM capabilities grow, their role in the arts is set to expand, and this study lays the groundwork for the automated evaluation of creative works at the intersection of AI and human creativity.

\section{Limitations} \label{section:Limitations}
While our study demonstrates the potential of LLMs for creative assessment, its conclusions are subject to several key limitations. First, our findings are constrained by the dataset, which is limited to 90 English-language poems. Generalizing these results will require validating our methodology on larger, multilingual corpora that encompass a greater diversity of poetic styles.

Second, our technical scope was narrow. The methodology, while inspired by CAT, requires direct comparison against expert human judges for full validation. Furthermore, our use of only two proprietary models, Claude-3-Opus and GPT-4o, means that the performance of other models, particularly open-source alternatives, remains an open question.

Finally, the broader ethical and social implications of using LLMs as evaluators must be addressed. Key risks include perpetuating algorithmic biases present in LLM training data and fostering creative homogenization, where creators may cater to machine preferences rather than pursuing novel expression. Navigating these challenges is essential for guiding these technologies toward enriching the creative landscape, not homogenizing it.

\section{Acknowledgments} 
We thank the anonymous reviewers for their valuable feedback.

The work of Dan Brown is supported by a Discovery Grant from the Natural Sciences and Engineering Council of Canada.

%\newpage

\bibliography{custom}

\clearpage
\appendix
\onecolumn

\section{Appendix} \label{section:appendix}

\subsection{LLM Prompts for In-Context Evaluation} \label{section:5_criteria_prompts}

All prompts for the in-context evaluations (Experiments 2 and 3) shared a common structure. We present the full prompt for the \textbf{Creativity} criterion below as a template. For the other four criteria, only the highlighted text was changed, as specified in the list that follows.

\vspace{5pt}
\hrule
\vspace{5pt}

\noindent\textbf{Full Prompt Template (using Creativity as an example):}

\begin{quote}
Below is the collection of 15 [or 90] poems. Evaluate the \textbf{creativity level} of each poem on the scale from 1 to 5, with 1 being \textbf{``least creative''} and 5 being \textbf{``most creative''}. Use the whole range of the scale, that is, the least creative poem in the collection must have the score of 1, and the most creative poem in the collection must have the score of 5. Use only whole integers without any decimal places.

Print out the filenames of the poems with their associated scores, ordered from the highest score to the lowest, in the following format:

\texttt{[position on the list]. [poems author] - [poems title] : [score]}

below are two example entries:

\texttt{1. Tom Smith - Some Poem : 5}

\texttt{2. Jane Jones - My Poem : 4}

\texttt{...}

POEMS:

===========================
\end{quote}

\vspace{5pt}
\hrule
\vspace{10pt}

\noindent\textbf{Variations for Other Criteria:}

\begin{description}
    \item[Quality:] Evaluate the \textbf{quality} of each poem [...] with 1 being \textbf{``lowest quality''} and 5 being \textbf{``highest quality''}.

    \item[Innovativeness:] Evaluate each text based on its \textbf{innovativeness} [...] with 1 indicating \textbf{``This poem is like other poems I have seen before''} and 5 indicating \textbf{``This poem is not like other poems I have seen before''}.

    \item[Similarity:] Evaluate each poem based on its \textbf{similarity to other poems you have read} [...] with 1 indicating \textbf{``not at all similar''} and 5 indicating \textbf{``highly similar''}.

    \item[Poeticness:] Evaluate each text based on its \textbf{qualification as a poem} [...] with 1 indicating \textbf{``this is not a poem''} and 5 indicating \textbf{``this is definitely a poem''}.
\end{description}

\clearpage

\subsection{Prompt for Without-Context Classification}
\label{section:GPT3-ABC}

The following prompt was used for the without-context classification task in Experiment 1. The model was instructed to insert the poem to be evaluated in place of the ellipsis.

\vspace{5pt}
\hrule
\vspace{5pt}

\begin{quote}
You will be evaluating a poem and categorizing it as ``Good'', ``Medium'', or ``Bad'' based on the following criteria:

\begin{itemize}
    \item \textbf{``Good''} poems are those that would be published in the foremost English-language poetry journal globally. These poems have been subjected to meticulous scrutiny by the magazine's editorial team.
    
    \item \textbf{``Medium''} poems are those that would be published in mid-level poetry magazines offering \$5-\$10 per poem. These poems have undergone some degree of editorial review.
    
    \item \textbf{``Bad''} poems are those that would be posted on a website for amateur poets and would receive no positive feedback. This website does not have any editorial filtration.
\end{itemize}

Here is the poem to evaluate:

\texttt{<poem>}

...

\texttt{</poem>}

Carefully read the poem and consider which category it belongs to based on the criteria above. Write your reasoning for the categorization inside \texttt{<reasoning>} tags. Then, output the final category you believe the poem belongs to inside \texttt{<category>} tags.
\end{quote}
\vspace{5pt}
\hrule
\vspace{10pt}

\subsection{Original Human Evaluation Criteria} \label{original criteria}

\begin{table*}[ht]
\centering % Changed from \begin{center}
\small
{

\begin{tabular}{lp{0.7\linewidth}} % Or adjust the width for the p column
\toprule
\textbf{Criterion} & \textbf{Description} \\
\midrule
Novelty & How different or unlike other poems the poem is. \\
Imagination & How imaginative or creative the author of the poem appears to be. \\
Value  & How good or worthwhile the poem is considered to be. \\
Quality & How good or high-quality the poem is as an example of poetry. \\
Appreciation & How well the author seems to understand how poetry works. \\
Skill  & How capable the author seems to be at writing poetry. \\
Typicality & How similar the poem is to other examples of poetry. \\
Wellbeing & How much the reader likes or enjoys the poem. \\
Effort  & How willing the reader is to spend time trying to understand the poem. \\
\bottomrule
\end{tabular}
}
\caption{The nine original evaluation criteria used by human non-expert judges in the study by Lamb et al. (2015). As discussed in Section~\ref{section:Methodology}, several of these criteria are inherently anthropocentric (e.g., ``Wellbeing,'' ``Effort'') or were found to be highly correlated, justifying our use of a more focused set of five criteria for the LLM evaluations.}\label{tbl:original_criteria}
\end{table*}

\clearpage

\subsection{Links to ``Good'' Poems from the Dataset} \label{llinks}

The following poems, categorized as ``Good'' in our dataset, are publicly available online through the Poetry Foundation. We provide these links for readers interested in examining examples of the source material.

\begin{itemize}
    \item A.E. Stallings, \href{https://www.poetryfoundation.org/poetrymagazine/poems/56772/the-companions-of-odysseus-in-hades}{\textit{The Companions of Odysseus in Hades}}
    \item Adam Fitzgerald, \href{https://www.poetryfoundation.org/poetrymagazine/poems/56685/poem-with-accidental-memory}{\textit{Poem with Accidental Memory}}
    \item Adrian Matejka, \href{https://www.poetryfoundation.org/poetrymagazine/poems/56681/gymnopedies-no-2}{\textit{Gymnopédies No. 2}}
    \item Aimee Nezhukumatathil, \href{https://www.poetryfoundation.org/poetrymagazine/poems/56555/two-moths}{\textit{Two Moths}}
    \item Danniel Schoonebeek, \href{https://www.poetryfoundation.org/poetrymagazine/poems/56697/a-woman-in-the-sun}{\textit{A Woman in the Sun}}
    \item Dolores Hayden, \href{https://www.poetryfoundation.org/poetrymagazine/poems/56882/flying-lesson}{\textit{Flying Lesson}}
    \item Franny Choi, \href{https://www.poetryfoundation.org/poetrymagazine/poems/56834/second-mouth}{\textit{Second Mouth}}
    \item Hannah Sanghee Park, \href{https://www.poetryfoundation.org/poetrymagazine/poems/56565/and-a-lie}{\textit{And a Lie}}
    \item Harmony Holiday, \href{https://www.poetryfoundation.org/poetrymagazine/poems/56577/gazelle-lost-in-watts}{\textit{Gazelle Lost in Watts}}
    \item Idra Novey, \href{https://www.poetryfoundation.org/poetrymagazine/poems/56683/the-duck-shit-at-clarion-creek}{\textit{The Duck Shit at Clarion Creek}}
    \item Ishion Hutchinson, \href{https://www.poetryfoundation.org/poetrymagazine/poems/56875/a-march}{\textit{A March}}
    \item K. Silem Mohammad, \href{https://www.poetryfoundation.org/poetrymagazine/poems/56767/from-the-sonnagrams}{\textit{from The Sonnagrams}}
    \item Karen An-Hwei Lee, \href{https://www.poetryfoundation.org/poetrymagazine/poems/56898/on-hierophany}{\textit{On Heirophany}}
    \item Kathleen Ossip, \href{https://www.poetryfoundation.org/poetrymagazine/poems/56631/elegies}{\textit{Elegies}}
    \item Laura Kasischke, \href{https://www.poetryfoundation.org/poetrymagazine/poems/56770/recall-the-carousel}{\textit{Recall the Carousel}}
    \item Leah Umansky, \href{https://www.poetryfoundation.org/poetrymagazine/poems/56676/i-want-to-be-starklike}{\textit{I Want to be Stark[like]}}
    \item Maria Melendez Kelson, \href{https://www.poetryfoundation.org/poetrymagazine/poems/56839/good-friday-56d239b7438cf}{\textit{Good Friday}}
    \item Marion McCready, \href{https://www.poetryfoundation.org/poetrymagazine/poems/56699/arrochar-alps}{\textit{Arrochar Alps}}
    \item Maxine Chernoff, \href{https://www.poetryfoundation.org/poetrymagazine/poems/56695/granted}{\textit{Granted}}
    \item Michael Earl Craig, \href{https://www.poetryfoundation.org/poetrymagazine/poems/56900/advice-for-horsemen}{\textit{Advice for Horsemen}}
    \item Michael Robbins, \href{https://www.poetryfoundation.org/poetrymagazine/poems/56691/sweet-virginia}{\textit{Sweet Virginia}}
    \item Myra Sklarew, \href{https://www.poetryfoundation.org/poetrymagazine/poems/56847/the-skin-of-sleep}{\textit{The Skin of Sleep}}
    \item Najwan Darwish, \href{https://www.poetryfoundation.org/poetrymagazine/poems/56902/mary-56d239d8bfc56}{\textit{Mary}} (trans. by Kareem James Abu-Zeid)
    \item Nance Van Winckel, \href{https://www.poetryfoundation.org/poetrymagazine/poems/56765/been-about}{\textit{Been About}}
    \item Natalie Shapero, \href{https://www.poetryfoundation.org/poetrymagazine/poems/56574/thirty-going}{\textit{Thirty Going}}
    \item Ocean Vuong, \href{https://www.poetryfoundation.org/poetrymagazine/poems/56768/detonation}{\textit{DetoNation}}
    \item Phillip B. Williams, \href{https://www.poetryfoundation.org/poetrymagazine/poems/56573/of-darker-ceremonies}{\textit{Of Darker Ceremonies}}
    \item Rae Armantrout, \href{https://www.poetryfoundation.org/poetrymagazine/poems/56650/geography-56d23958d6ab7}{\textit{Geography}}
    \item Rebecca Hazelton, \href{https://www.poetryfoundation.org/poetrymagazine/poems/56633/trying-fourleggedness}{\textit{Trying Fourleggedness}}
    \item Samiya Bashir, \href{https://www.poetryfoundation.org/poetrymagazine/poems/56890/carnot-cycle}{\textit{Carnot Cycle}}
\end{itemize}

 \subsection{Visual Representation of Poem Orderings}\label{app:Visual Representation of Poem Orderings}

\begin{figure}[H]

\begin{center}

% Adjust the rectangles' width and height as needed
\newcommand{\rectWidth}{0.08} % Width in cm
\newcommand{\rectHeight}{0.28} % Height in cm
\small
\setlength{\tabcolsep}{4pt}
\begin{tabular}{@{}lcr@{}}
\toprule
\textbf{Model/Criterion} & \textbf{Ordering} & \textbf{SRC} \\
\bottomrule
\end{tabular}%

\vspace{2mm}
\end{center}
\caption{Comprehensive visual summary of all poem orderings from the LLM evaluations in Experiments 2 and 3, across all tested criteria and methods. Each poem is colour-coded by its ground truth category and ranked from highest (left) to lowest (right). The lower section includes the human non-expert evaluations from~\citet{lamb2015human} to provide a direct visual baseline for performance comparison.}
\label{fig:ordering_all4_human}

\end{figure}

\clearpage

\subsection{Detailed ANOVA Results}\label{ANOVA results for poem evaluation in five criteria in Experiments 2 and 3}

\begin{table*}[htpb]
\centering % Changed from \begin{center}
\small
\setlength{\tabcolsep}{4pt} % Retained custom tabcolsep
{
% l for Criterion, r for all numerical/scientific notation columns
\begin{tabular}{l cccrr} 
\toprule
\textbf{Criterion} & \textbf{Good} & \textbf{Medium} & \textbf{Bad} & \textbf{F-value} & \textbf{p-value} \\
\midrule
\multicolumn{6}{c}{\textbf{GPT-4o}} \\ % Spanning all 6 columns now
\cmidrule(lr){1-6} % Rule under the multicolumn, spanning all 6 columns
Average Innovativeness Score & 3.49  & 3.12  & 1.64 & 82.72  & 7.5e-21  \\
Average Quality Score        & 3.48  & 2.98  & 1.71 & 92.31  & 3.1e-22  \\
Average Creativity Score     & 3.46  & 3.15  & 1.77 & 78.8   & 2.96e-20 \\
Average Poeticness Score     & 3.82  & 3.42  & 2.15 & 60.27  & 3.75e-17 \\
Average Similarity Score     & 3.41  & 2.86  & 2.13 & 49.28  & 4.9e-15  \\
\midrule % Separator for the next section
\multicolumn{6}{c}{\textbf{GPT-4o 15n}} \\
\cmidrule(lr){1-6}
Average Innovativeness Score & 10.99 & 9.57  & 3.63 & 108.06 & 2.62e-24 \\
Average Quality Score        & 11.32 & 9.30  & 3.50 & 157.45 & 1.23e-29 \\
Average Creativity Score     & 10.81 & 9.63  & 3.79 & 96.72  & 7.73e-23 \\
Average Poeticness Score     & 10.72 & 9.11  & 4.14 & 71.43  & 4.42e-19 \\
Average Similarity Score     & 10.35 & 8.29  & 5.28 & 57.56  & 1.19e-16 \\
\midrule
\multicolumn{6}{c}{\textbf{Claude-3-Opus}} \\
\cmidrule(lr){1-6}
Average Innovativeness Score & 3.24  & 2.77  & 1.54 & 36.3   & 3.44e-12 \\
Average Quality Score        & 3.52  & 2.84  & 1.43 & 94.47  & 1.56e-22 \\
Average Creativity Score     & 3.49  & 2.93  & 1.64 & 50.04  & 3.43e-15 \\
Average Poeticness Score     & 3.85  & 3.2   & 1.8  & 72.02  & 3.53e-19 \\
Average Similarity Score     & 3.37  & 2.69  & 1.58 & 90.5   & 5.55e-22 \\
\midrule
\multicolumn{6}{c}{\textbf{Claude-3-Opus 15n}} \\
\cmidrule(lr){1-6}
Average Innovativeness Score & 10.83 & 9.02  & 4.26 & 52.89  & 9.3e-16  \\
Average Quality Score        & 11.65 & 9.06  & 3.33 & 132.23 & 4.2e-27  \\
Average Creativity Score     & 11.11 & 8.95  & 4.02 & 62.55  & 1.46e-17 \\
Average Poeticness Score     & 11.42 & 8.81  & 3.79 & 85.15  & 3.27e-21 \\
Average Similarity Score     & 11.22 & 8.63  & 4.13 & 111.08 & 1.11e-24 \\
\midrule
\multicolumn{6}{c}{\textbf{Claude-3-Opus 90n}} \\
\cmidrule(lr){1-6}
Average Innovativeness Score & 56.5  & 48.9  & 31.1 & 41.2   & 2.64e-13 \\
Average Quality Score        & 53.7  & 50.9  & 32.9 & 25.74  & 1.66e-9  \\
Average Creativity Score     & 54.6  & 48.1  & 33.8 & 31.22  & 6.02e-11 \\
Average Poeticness Score     & 56.2  & 48.6  & 31.7 & 35.9   & 4.29e-12 \\
Average Similarity Score     & 51.2  & 47.8  & 37.5 & 16.13  & 1.1e-6   \\
\midrule
\multicolumn{6}{c}{\textbf{Claude-3-Opus 90}} \\
\cmidrule(lr){1-6}
Average Innovativeness Score & 3.04  & 2.82  & 2.26 & 30.09  & 1.96e-12 \\
Average Quality Score        & 2.6   & 2.55  & 1.92 & 24.57  & 3.47e-9  \\
Average Creativity Score     & 3.13  & 2.91  & 2.51 & 19.84  & 7.94e-8  \\
Average Poeticness Score     & 2.94  & 2.67  & 1.97 & 37.34  & 1.96e-12 \\
Average Similarity Score     & 2.56  & 2.41  & 2.11 & 14.95  & 2.63e-6  \\
\bottomrule
\end{tabular}
}
\caption{Detailed ANOVA results, confirming that all LLM evaluation methods significantly distinguish between the ``Good,'' ``Medium,'' and ``Bad'' poem categories. The high F-values indicate a large variance between the group means, and the extremely low p-values (p << 0.01) confirm that these differences are statistically significant across all criteria. The ``Good,'' ``Medium,'' and ``Bad''columns show the mean scores assigned to poems in each ground truth category. Notably, the rank-based scoring methods (e.g., ``15n'') consistently yield higher F-values, indicating a stronger separation between the categories.}\label{tbl:var_1_results_full}

\end{table*}

\clearpage % Start a fresh page
\onecolumn  % Switch to a one-column layout for the appendix page

\subsection{Illustrating LLM Output Variability and Consistency}\label{app:repeated_evaluations_example}

\begin{table}[H]
\centering
% \small % Kept commented as per original
% l for Poem ID/Author - Poem, c for Category, r for all score columns and Average Score
\begin{tabular}{l c rrrrrrrrrr r} 
\toprule
\multicolumn{13}{c}{\textbf{Claude-3-Opus 15n}} \\
\cmidrule(lr){1-13} % Light rule under section header
\textbf{Poem ID} & \textbf{Cat.} & \multicolumn{10}{c}{\textbf{Scores}} & \textbf{Average Score} \\
\cmidrule(lr){3-12} % Rule under "Scores" multicolumn, spanning cols 3-12
\midrule % Heavier rule separating headers from data
Poem 27 & A & 14 & 15 & 15 & 14 & 15 & 15 & 14 & 15 & 15 & 14 & 14.6 \\
Poem 3  & A & 15 & 13 & 12 & 13 & 14 & 10 & 15 & 11 & 13 & 11 & 12.7 \\
Poem 7  & A & 13 & 14 & 9  & 11 & 13 & 14 & 13 & 14 & 14 & 12 & 12.7 \\
Poem 6  & A & 12 & 12 & 13 & 15 & 12 & 11 & 8  & 12 & 11 & 15 & 12.1 \\
Poem 8  & A & 11 & 9  & 14 & 10 & 10 & 13 & 7  & 13 & 12 & 8  & 10.7 \\
Poem 50 & B & 9  & 7  & 10 & 12 & 9  & 12 & 11 & 8  & 8  & 9  & 9.5  \\
Poem 54 & B & 7  & 11 & 7  & 6  & 11 & 9  & 12 & 9  & 9  & 13 & 9.4  \\
Poem 53 & B & 10 & 10 & 11 & 9  & 8  & 8  & 10 & 7  & 10 & 10 & 9.3  \\
Poem 41 & B & 8  & 8  & 8  & 8  & 7  & 7  & 6  & 10 & 7  & 6  & 7.5  \\
Poem 42 & B & 6  & 6  & 5  & 7  & 6  & 6  & 9  & 5  & 5  & 5  & 6.0  \\
Poem 61 & C & 5  & 5  & 6  & 5  & 5  & 5  & 5  & 6  & 6  & 7  & 5.5  \\
Poem 74 & C & 4  & 3  & 3  & 4  & 4  & 2  & 4  & 4  & 4  & 4  & 3.6  \\
Poem 79 & C & 3  & 2  & 2  & 3  & 3  & 1  & 3  & 2  & 3  & 1  & 2.3  \\
Poem 65 & C & 2  & 1  & 4  & 2  & 2  & 3  & 1  & 3  & 2  & 3  & 2.3  \\
Poem 69 & C & 1  & 4  & 1  & 1  & 1  & 4  & 2  & 1  & 1  & 2  & 1.8  \\
\midrule % Separator between the two sections
\multicolumn{13}{c}{\textbf{Claude-3-Opus 15}} \\
\cmidrule(lr){1-13}
% Note: Header for first column is different here in the original.
\textbf{Poem ID} & \textbf{Cat.} & \multicolumn{10}{c}{\textbf{Scores}} & \textbf{Average Score} \\ 
\cmidrule(lr){3-12}
\midrule
Poem 27 & A & 4  & 5  & 5  & 4  & 5  & 5  & 4  & 5  & 5  & 4  & 4.6  \\
Poem 3  & A & 5  & 3  & 4  & 4  & 4  & 3  & 5  & 3  & 4  & 3  & 3.8  \\
Poem 7  & A & 4  & 4  & 3  & 3  & 4  & 4  & 4  & 4  & 4  & 3  & 3.7  \\
Poem 6  & A & 3  & 2  & 4  & 5  & 3  & 3  & 3  & 3  & 3  & 5  & 3.4  \\
Poem 50 & B & 3  & 1  & 4  & 4  & 3  & 4  & 3  & 3  & 3  & 3  & 3.1  \\
Poem 53 & B & 3  & 1  & 4  & 3  & 3  & 3  & 3  & 3  & 3  & 3  & 2.9  \\
Poem 54 & B & 3  & 1  & 3  & 2  & 3  & 3  & 3  & 3  & 3  & 4  & 2.8  \\
Poem 8  & A & 3  & 1  & 4  & 3  & 3  & 4  & 2  & 3  & 3  & 2  & 2.8  \\
Poem 41 & B & 3  & 1  & 3  & 3  & 3  & 2  & 2  & 3  & 2  & 2  & 2.4  \\
Poem 42 & B & 2  & 1  & 2  & 2  & 2  & 2  & 3  & 3  & 2  & 2  & 2.1  \\
Poem 61 & C & 2  & 1  & 3  & 2  & 2  & 2  & 2  & 3  & 2  & 2  & 2.1  \\
Poem 74 & C & 2  & 1  & 2  & 2  & 2  & 2  & 2  & 2  & 2  & 2  & 1.9  \\
Poem 65 & C & 2  & 1  & 2  & 1  & 2  & 2  & 1  & 2  & 1  & 1  & 1.5  \\
Poem 79 & C & 2  & 1  & 2  & 1  & 2  & 1  & 1  & 2  & 1  & 1  & 1.4  \\
Poem 69 & C & 1  & 1  & 1  & 1  & 1  & 2  & 1  & 1  & 1  & 1  & 1.1  \\
\bottomrule
\end{tabular}
\caption{Sample output from the inter-rater reliability test (Experiment 4), showing 10 repeated evaluations of a single 15-poem subset for the ``Creativity'' criterion. Results are shown for both the rank-based scoring (Claude-3-Opus 15n) and the raw 1-5 scoring (Claude-3-Opus). The table illustrates that while temperature=1 introduces score variations in each run, the relative ordering of poems remains largely consistent, forming the basis for the high Intraclass Correlation Coefficient (ICC) values reported in Table~\ref{tab:ICC_full}. Poem numbers correspond to their ground truth category: 1-30 (A: Good), 31-60 (B: Medium), 61-90 (C: Bad).}
\label{tab:poetry_scores_example_iccc}
\end{table}

\clearpage

\subsection{Results of ICC evaluations}\label{app:Results of ICC evaluations}

\begin{table}[htbp] % Added placement specifier
\centering
\footnotesize % Retained scriptsize
\begin{tabular}{l rrr rrr rrr} % l for first col, r for all numerical/sci-notation
\toprule
\multirow{2}{*}{\textbf{Criteria \& Model}} & \multicolumn{3}{c}{ICC1k} & \multicolumn{3}{c}{ICC2k} & \multicolumn{3}{c}{ICC3k} \\
\cmidrule(lr){2-4} \cmidrule(lr){5-7} \cmidrule(lr){8-10} % cmidrules for sub-headers
 & \textbf{ICC} & \textbf{F} & \textbf{p-value} & \textbf{ICC} & \textbf{F} & \textbf{p-value} & \textbf{ICC} & \textbf{F} & \textbf{p-value} \\
\midrule
\multicolumn{10}{c}{\textbf{CREATIVITY}} \\
\cmidrule(lr){1-10} % Light rule under section header
Claude-3-Opus     & 0.98 & 54.66 & 1.63e-48 & 0.98 & 53.59 & 4.14e-46 & 0.98 & 53.59 & 4.14e-46 \\
Claude-3-Opus 15n & 0.99 & 80.84 & 2.12e-58 & 0.99 & 75.45 & 3.4e-54  & 0.99 & 75.45 & 3.4e-54  \\
GPT-4o            & 0.97 & 36.25 & 8.17e-39 & 0.97 & 42.59 & 6.54e-41 & 0.98 & 42.59 & 6.54e-41 \\
GPT-4o 15n        & 0.99 & 92.32 & 7.22e-62 & 0.99 & 86.17 & 1.98e-57 & 0.99 & 86.17 & 1.98e-57 \\ 
\addlinespace % Replaces the \\ \\ for better vertical spacing between sections
\multicolumn{10}{c}{\textbf{INNOVATIVENESS}} \\
\cmidrule(lr){1-10}
Claude-3-Opus     & 0.99 & 101.03 & 2.99e-64 & 0.99 & 105.98 & 1.47e-62 & 0.99 & 105.98 & 1.47e-62 \\
Claude-3-Opus 15n & 0.99 & 161.46 & 6.16e-77 & 0.99 & 150.7  & 1.74e-71 & 0.99 & 150.7  & 1.74e-71 \\
GPT-4o            & 0.98 & 44.61  & 1.28e-43 & 0.98 & 51.25  & 4.41e-45 & 0.98 & 51.25  & 4.41e-45 \\
GPT-4o 15n        & 0.99 & 75.05  & 1.75e-56 & 0.99 & 70.05  & 2.09e-52 & 0.99 & 70.05  & 2.09e-52 \\ 
\addlinespace
\multicolumn{10}{c}{\textbf{POETICNESS}} \\
\cmidrule(lr){1-10}
Claude-3-Opus     & 0.99 & 98.95  & 1.07e-63 & 0.99 & 102.67 & 9.13e-62 & 0.99 & 102.67 & 9.13e-62 \\
Claude-3-Opus 15n & 0.99 & 147.33 & 2e-74    & 0.99 & 137.51 & 3.83e-69 & 0.99 & 137.51 & 3.83e-69 \\
GPT-4o            & 0.98 & 43.34  & 6.17e-43 & 0.98 & 49.62  & 2.4e-44  & 0.98 & 49.62  & 2.4e-44  \\
GPT-4o 15n        & 0.99 & 94.85  & 1.4e-62  & 0.99 & 88.52  & 4.31e-58 & 0.99 & 88.52  & 4.31e-58 \\ 
\addlinespace
\multicolumn{10}{c}{\textbf{QUALITY}} \\
\cmidrule(lr){1-10}
Claude-3-Opus     & 0.99 & 93.41  & 3.56e-62 & 0.99 & 98.77  & 8.39e-61 & 0.99 & 98.77  & 8.39e-61 \\
Claude-3-Opus 15n & 0.99 & 156.2  & 5e-76    & 0.99 & 145.79 & 1.23e-70 & 0.99 & 145.79 & 1.23e-70 \\
GPT-4o            & 0.95 & 21.74  & 5.24e-28 & 0.95 & 29.67  & 3.28e-33 & 0.97 & 29.67  & 3.28e-33 \\
GPT-4o 15n        & 0.99 & 83.52  & 3e-59    & 0.99 & 77.96  & 5.5e-55  & 0.99 & 77.96  & 5.5e-55  \\ 
\addlinespace
\multicolumn{10}{c}{\textbf{SIMILARITY}} \\
\cmidrule(lr){1-10}
Claude-3-Opus     & 0.94 & 15.46  & 9.2e-22  & 0.94 & 15.66  & 2.2e-21  & 0.94 & 15.66  & 2.2e-21  \\
Claude-3-Opus 15n & 0.96 & 25.74  & 2.1e-31  & 0.96 & 24.03  & 4.93e-29 & 0.96 & 24.03  & 4.93e-29 \\
GPT-4o            & 0.90  & 10.37  & 1.46e-15 & 0.90  & 9.90    & 1.35e-14 & 0.90  & 9.90    & 1.35e-14 \\
GPT-4o 15n        & 0.92 & 12.06  & 9.27e-18 & 0.92 & 11.26  & 2.42e-16 & 0.91 & 11.26  & 2.42e-16 \\ 
\bottomrule
\end{tabular}
\caption{Intraclass Correlation Coefficient (ICC) results from Experiment 4, measuring the inter-rater reliability of each LLM method across 10 repeated runs. The exceptionally high ICC values (typically > 0.90) and statistically significant p-values demonstrate a very high degree of consistency. These findings confirm that the LLM evaluations are robust and reliable, even with the variability introduced by setting temperature=1.}
\label{tab:ICC_full}
\end{table}
\clearpage

\end{document}